# High-Speed Cornering Control and Real-Vehicle Deployment for Autonomous Electric Vehicles

Shiyue Zhao, Junzhi Zhang, Neda Masoud, Yuhong Jiang, Heye Huang and Tao Liu

*Abstract*—Executing drift maneuvers during high-speed cornering presents significant challenges for autonomous vehicles, yet offers the potential to minimize turning time and enhance driving dynamics. While reinforcement learning (RL) has shown promising results in simulated environments, discrepancies between simulations and real-world conditions have limited its practical deployment. This study introduces an innovative control framework that integrates trajectory optimization with drift maneuvers, aiming to improve the algorithm's adaptability for real-vehicle implementation. We leveraged Bezier-based pre-trajectory optimization to enhance rewards and optimize the controller through Twin Delayed Deep Deterministic Policy Gradient (TD3) in a simulated environment. For real-world deployment, we implement a hybrid RL-MPC fusion mechanism, , where TD3-derived maneuvers serve as primary inputs for a Model Predictive Controller (MPC). This integration enables precise real-time tracking of the optimal trajectory, with MPC providing corrective inputs to bridge the gap between simulation and reality. The efficacy of this method is validated through real-vehicle tests on consumer-grade electric vehicles, focusing on drift U-turns and drift right-angle turns. The control outcomes of these real-vehicle tests are thoroughly documented in the paper, supported by supplementary video evidence. Notably, this study is the first to deploy and apply an RL-based transient drift cornering algorithm on consumer-grade electric vehicles.

## I. INTRODUCTION

The rapid evolution of vehicle intelligence and networked technologies has propelled autonomous vehicles towards achieving L4 and L5 levels of automation [1,2]. This technological leap has sparked considerable interest in autonomous sports cars, which traditionally referred to high-performance vehicles in racing but now include a wide range of high-performance automobiles [3,4]. Sports cars, to maximize driving pleasure and achieve quicker lap times, require exceptional acceleration capabilities. Expert drivers often leverage rear-wheel saturation and significant lateral displacement of the vehicle's center of mass to execute drifts, a technique that exceeds standard driving limits [5].

Drift cornering represents one of the most extreme and challenging tasks in autonomous driving. This maneuver pushes the vehicle's tires into highly saturated, nonlinear states, requiring advanced control strategies to manage the complex dynamics involved [6]. The ability to drift effectively hinges on the precise manipulation of the vehicle's posture, especially during sharp turns, where rapid adjustments are essential to handle transient drift states [7]. This area of research is critical not only for enhancing performance but also for ensuring the stability and safety of autonomous sports cars under extreme driving conditions.

Numerous studies have explored the mechanics and control strategies for drift cornering. Cai et al. [8] proposed a deep reinforcement learning algorithm to control an unmanned race car during high-speed cornering. After training on tracks of varying difficulty, the controller could make the vehicle drift smoothly and quickly through sharp turns on unknown maps. Hou et al. [9] used expert example data to establish basic maneuvers and then applied a data-driven TD3 (Temporal Difference) reinforcement learning algorithm to generate residual terms, thereby enhancing cornering speeds. M. Liu et al. [10] introduced a segmented drift parking methodology employing a model predictive controller for the approach phase and an open-loop control law for the drift phase. G. Chen et al. [11] proposed a hierarchical dynamic drift controller (HDDC) consisting of three layers for drifting maneuvers and typical turning maneuvers to obtain practical path-following control. These studies have achieved promising results in understanding drift cornering mechanics and designing control methods. However, they have primarily focused on simulation environments where conditions are controlled and predictable [12]. While these approaches have demonstrated satisfactory results in simulation environments, their applicability to real vehicle traffic environments remains to be further validated.

Deploying drift cornering control strategies in actual vehicles and traffic environments introduces additional complexities. Real-world conditions differ significantly from controlled simulation environments, introducing unpredictability that can impact performance and safety [13,14]. Some studies have attempted to deploy designed drift cornering controllers in real vehicles. For instance, J. Christian Gerde et al. [15,16] achieved steady-state drifting and figure-eight maneuvers in real vehicles using feedforward-feedback controllers and nonlinear model predictive control (NMPC). F. Zhang et al. [17], inspired by professional drivers, proposed a rule-based algorithm to plan reference drift trajectories with high side-slip angles along sharp curves and conducted tests using a 1/10 scale radio-controlled (RC) vehicle. S. Zhao et al. [18] developed an adaptive drift controller based on a multilayer neural network and the fast start soft actor-critic algorithm, performing short-term tests on a highly controllable 1/2 scale PIX chassis.

In summary, most current drift strategies deployed in real vehicles rely on a hierarchical trajectory planning-tracking architecture. While these strategies can effectively track target drift dynamics, they struggle to optimize for minimal cornering time on complex curves. [19,20]. Currently, most decision-making methods aimed at optimizing cornering time are based on scenario-to-action reinforcement learning (RL), which involves transient drift control with continuous state changes during cornering. However, RL-based controllers face significant simulation-to-reality discrepancies. [8,21-22]. Specifically, RL controllers are trained using interactive data in simulation environments, where the modeled dynamics and road conditions often fail to fully capture the complexity and variability of real vehicles and scenarios, leading to challenges when deployed in real-world vehicles. [23-24].

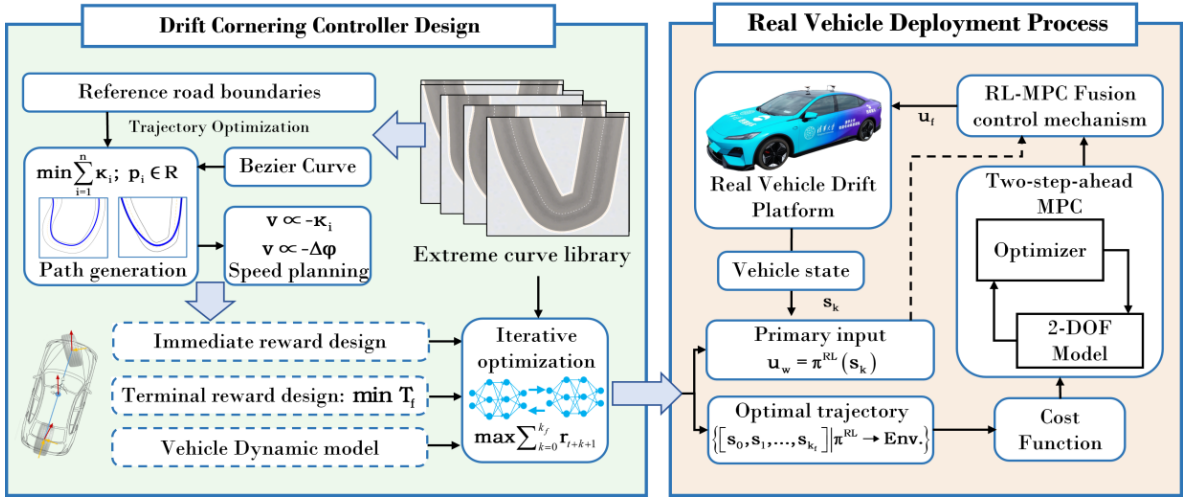

**Fig. 1.** Drift Cornering Controller Design and Real Vehicle Deployment Process Overview. The left side outlines the controller design, featuring trajectory optimization with Bezier curves, velocity planning, and reward-influenced controllers via reinforcement learning.

To address the simulation-to-deployment challenges of an RL-based high-speed cornering controller, we propose a control framework that integrates data-driven and model-based tracking strategies. Our main contributions are as follows:

1) Proposed a TD3-based controller combined with pre-trajectory optimization, capable of performing transient drift cornering aimed at optimizing cornering time on complex curves.
2) Designed a hybrid control framework integrating RL and MPC, improving adaptability for real-world vehicle deployment and addressing simulation-to-reality gaps.
3) Validated the proposed control framework on **consumer-grade electric vehicles,** successfully executing U-turns and right-angle turns, demonstrating the application of the proposed controller in unpredictable environments.

**Notably, this study is the first to deploy and apply a scenario-to-action RL-based transient drift cornering algorithm on consumer-grade electric vehicles.** Regarding practical applications, this work significantly advances autonomous racing by reducing cornering time. It also enhances the handling of consumer-grade vehicles during high-speed cornering, enabling high-speed drift maneuvers. Additionally, by effectively bridging simulation-to-reality gaps, we provide a viable solution for deploying scenario-to-action RL controllers in real-world vehicles, promoting the practical application of reinforcement learning in unpredictable environments.

II. PROBLEM DEFINITION AND OVERALL IMPLEMENTATION

Drifting around extreme corners is crucial for professional racers aiming to minimize overall lap times. For autonomous vehicles, effectively controlling drift on these challenging bends and deploying these capabilities in real vehicles present significant challenges. To address this challenge, we proposed a solution architecture as illustrated in Figure 1. The diagram is divided into two main parts: controller design on the left and the application process for real-vehicle deployment process on the right.

First, we gathered typical extreme curves from racing tracks to compile an extensive limit curve library for this study, significantly enhancing our analysis and implementation of drift control strategies. The iterative process of RL heavily depends on the reward function design, with the main reward for extreme cornering being the terminal reward—specifically, the reduction of cornering time. Relying solely on this reward configuration for iterative optimization can lead to severe issues with sparse rewards. To mitigate this, we developed a pre-trajectory optimization method that utilizes Bezier curves for path optimization based on the principle of minimum curvature. This method also generates speed plans based on curvature and heading angle deviations. During the reinforcement learning iterative optimization process, this pre-optimized trajectory is used to define immediate rewards. Furthermore, the iterative optimization of the controller is facilitated through its interaction with the simulated environment. In this setup, the controller is a deep neural network that processes the current environmental state as its input and outputs specific values for continuous actions.

Network-based controllers perform well in training environments but struggle in unfamiliar scenarios, especially when moving from simulation to real vehicles. This is a significant challenge for reinforcement learning-based controllers in actual road conditions. To address this, we explore scenario-based vehicle deployment, as shown in Figure 1.

After establishing target curve information, a network-based controller is deployed in a virtual environment to derive the optimal trajectory. These inputs are crucial for executing drift maneuvers in nonlinear regimes. To apply this in real vehicles, we developed an RL-MPC fusion control mechanism. The network-based controller's inputs act as primary inputs, enhancing initial response, while a two-step forward model predictive controller (MPC) precisely follows the trajectory. This hybrid strategy ensures robust handling and accurate trajectory tracking, bridging the gap between simulation and real-world application.

### III. DRIFT CORNERING CONTROLLER DESIGN

This section delves into the detailed design and optimization of the network-based drift cornering controller.

The primary objective of the pre-trajectory is to provide a reliable path for the vehicle, guiding the agent to successfully complete the cornering task in the early stages of training. While any feasible trajectory could serve this purpose, Bézier curves offer a closer approximation to the optimal solution [25], thereby improving exploration efficiency in the later stages of training. This is crucial because, as training progresses, the influence of the terminal reward increases, and the impact of following the pre-optimized trajectory diminishes, shifting the focus toward minimizing cornering time.

*A. Pre-trajectory optimization*

In this study, we selected Bézier-based pre-trajectory optimization over alternative approaches due to its superior ability to generate smooth and feasible trajectories that effectively guide the vehicle during the initial phases of RL training [25]. Firstly, we transform the curve coordinates $[x, y]$ into a *Frenet* coordinates $[s, l]$, incorporating its intrinsic radian information, as shown below:

$$s = \underset{s^* \in [0, s_0]}{argmin} \left| l\left(s^*\right) \right| \quad (1)$$

$$l = sign\left[\left(y - y(s)\right) \cdot \cos\alpha - \left(x - x(s)\right) \cdot \sin\alpha\right] \cdot \left|b(s)\right| \quad (2)$$

where the symbol $s$ represents the distance moved along the track's tangent direction, while $l$ denotes the lateral distance from the track centerline. The calculation method involves fitting a polynomial to the function $b(s)$, which is done using multiple adjacent samples along the curve.

Following this transformation, the track is discretized within this coordinate system to obtain discrete road points $P_d$.

$$P_d = \left\{ \left( i \cdot i, l_{\min} + j \frac{l_{\max} - l_{\min}}{m} \right) \middle| i = [0, n]; j = [0, m-1] \right\} \quad (3)$$

The path is discretized by taking points at $s_{max}/n$ meter intervals along the $s$-axis, with each s value corresponding to $n$ evenly spaced points along the $l$-axis. These points, which represent the spatial configuration of the track, are illustrated in Figure 2.

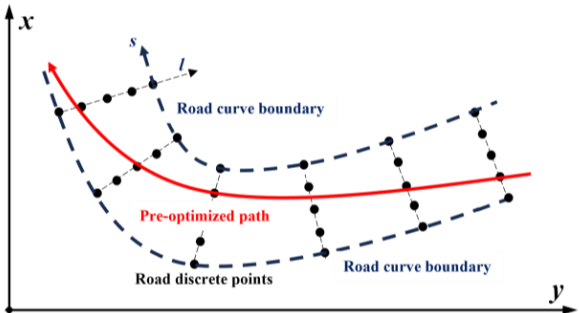

**Fig. 2.** Path Planning in Pre-Trajectory Optimization

Furthermore, we select reference points for path generation based on the minimum curvature principle. Minimizing curvature allows the vehicle to maintain higher speeds along the path, reducing cornering time. Simulations show that minimizing the curvature integral often results in shorter cornering times than minimizing path length, as smoother paths enable better stability and speed control throughout the trajectory. A cubic polynomial function $l(s)$ is fitted to represent the lateral offset of the path in *Frenet* coordinates, as shown in Equation 4. This method ensures a smooth and continuous representation of the path's deviation from the centerline.

$$l(s) = a_0 + a_1 \cdot s + a_2 \cdot s^2 + a_3 \cdot s^3 \quad (4)$$

where the coefficients $[a_0, a_1, a_2, a_3]$ are determined by solving a least-squares optimization problem using the discrete road points. The curvature $\kappa(s)$ of the path can be calculated using the following equation:

$$\kappa(s) = \frac{d^2 l(s)/ds^2}{\left(1 + \left(dl(s)/ds\right)^2\right)^{3/2}} \quad (5)$$

To find the path with the minimum curvature, we set up an optimization problem. The objective function $J$ to minimize is typically the integral of the squared curvature over the length of the path:

$$J = \int_0^{s_{\max}} \kappa(s)^2 \, ds \quad (6)$$

$$s.t. \quad l(0)' = l_0', \quad l(s_{\max})' = l_{\max}'$$

The path generation is constrained by requirements that it must meet boundary conditions, including the initial and final orientations of the path. This optimization problem is typically solved numerically, often employing methods like gradient descent. By adhering to these procedures, smooth paths with minimal curvature are generated, ensuring that the vehicle follows a safe and efficient path around sharp bends. Once this optimal path is generated, vehicle speed is planned based on the curvature of the path to ensure stability and control. Vehicle speed planning takes into account the stability constraint $|\dot{\varphi} \cdot v| \leq |\mu \cdot g|$, combined with $\dot{\varphi} = v \cdot \kappa$, yielding the following Equation:

$$v_d \leq \sqrt{\left|\frac{\mu g}{\kappa(s)}\right|} \quad (7)$$

where $\dot{\varphi}$ represents the yaw angular velocity of the vehicle, $v_d$ refers to the expected speed plan of the vehicle, and $\mu$ denotes the ground adhesion coefficient.

It is noteworthy that while the generated trajectories ensure feasible cornering paths, pre-optimized trajectories closer to the optimal solution enhance exploration efficiency in later training stages. We compared Bézier curve-based optimization with using the road centerline for RL trajectory tracking. Under identical conditions of 5000 training episodes and the same starting points, the cornering times for 90-degree turns, 135-degree turns, and U-turns are shown in Table I.

TABLE I COMPARISON OF CORNERING TIMES USING ROAD CENTERLINE VS. BÉZIER CURVE OPTIMIZATION

| Scenario | Road Centerline (s) | Bézier Optimization (s) |
| --- | --- | --- |
| 90-degree turn | 2.94 | **2.81** |
| 135-degree turn | 4.59 | **4.09** |
| U-turn | 4.96 | **4.52** |

*B. Reward design*

The reward function plays a crucial role in guiding the iterative optimization process of reinforcement learning, which is expressed as a function:

$$r = r_i + r_t = r_p + r_s + r_m + r_t \quad (8)$$

where $r_i$ represents the immediate rewards, while $r_t$ denotes the terminal rewards.

The design of immediate rewards focuses on two primary objectives. The first and most critical objective is to ensure rigorous adherence to the pre-optimized trajectory. The second objective is to encourage extreme driving with substantial side-slip angles in sharp corners. The detailed calculation of each reward component is as follows:

$$r_p = k_{pl} \cdot |l - l_{ref}(s)| + k_{pv} \cdot |v - v_d^{max}| \quad (9)$$

where $l(s)$ represents the lateral offset of the pre-optimized trajectory corresponding to the current $s$ coordinate, and $v_d^{max}$ is the maximum vehicle speed that satisfies the stability constraints under the current path, as shown in Equation 7. The constants $k_{pl}$ and $k_{pv}$ are negative values used to adjust the weight for tracking the rewards of the pre-optimized trajectory. The reward component $r_p$ encourages driving along the pre-optimized path at the maximum speed permitted by the stability constraints, thereby enhancing the exploration efficiency during iterative optimization.

The side-slip angle, a key indicator for drift intensity, is measured using the rear axle center as a reference point ($b_r = 0$), eliminating the impact of the kinematic side-slip angle ($\beta_{kin}$).

$$\beta_r = arctan\left(\frac{v_y}{v_x}\right) \quad (10)$$

$v_x$ corresponds to the vehicle's longitudinal speed, while $v_y$ indicates its sideways velocity, both measured in the vehicle's local coordinates.

$$r_s = k_s \cdot \left(1 - e^{k_{s1} \cdot \beta_r}\right) \quad (11)$$

where $\beta$ is the center of mass side-slip angle of the vehicle, $k_s$ is a positive constant, and $k_{s1}$ is a negative constant. The reward component $r_s$ incentivizes the vehicle to navigate corners with a high center-of-mass side-slip angle. In high-speed racing, drivers use transient drifts with large side-slip angles to minimize cornering time. Prior knowledge and training experience show that rewarding $r_s$ helps guide the agent toward a global optimum in later training stages. To ensure the smoothness, we set $r_m$ to penalize high rates of change in the input, which will not be explained in detail here.

The terminal reward $r_t$ is designed to reflect the controller's performance upon task completion. Specifically, the $r_t$ is based on the total time required to complete the track, serving as a direct measure of driving performance and efficiency. As the agent successfully completes the entire cornering task, the weight of the reward associated with the total cornering time gradually increases, diminishing the impact of rewards tied to tracking the pre-optimized trajectory.

$$r_t = k_{t1} \cdot s + k_{t2} \cdot \chi \cdot (t_f - t_{ref}) \quad (12)$$

where $\chi$ represents the parameter indicating the final state of the vehicle: $\chi = 1$ signifies that the vehicle has completed the cornering task safely, while $\chi = 0$ indicates an unsafe event such as a collision with the track boundary or a rollover. Also, $t_f$ represents the time taken by the vehicle to reach the terminal state, and $t_{ref}$ is the total driving time of the pre-optimized trajectory. The constant $k_{t1}$ is a positive value designed to reward the vehicle for approaching the cornering endpoint as closely as possible during the early stages of training, while still ensuring the completion of the entire cornering task when optimizing for cornering time in later stages. The constant $k_{t1}$, on the other hand, is a positive value that incentivizes achieving the shortest possible extreme cornering time.

*C. Iterative Training Setup*

The state space defines the vehicle's condition and are crucial for the control and optimization process, which is defined as:

$$S = [s, l, \alpha, \dot{s}, \dot{l}, \dot{\alpha}, v_x, v_y, \chi, \kappa]$$

Here, $s$ and $l$ are the position coordinates in the *Frenet* coordinate system, detailed in subsection III.A. $\kappa$ contains the curvature information of the track at a certain distance ahead. The attitude angle $\alpha$ in the *Frenet* coordinate system is calculated as follows:

$$\alpha = \varphi - \varphi_{ref}(s) \quad (13)$$

where $\varphi_{ref}(s)$ represents the track angle in the Cartesian coordinate system corresponding to the current position of the vehicle. Additionally, $s'$ and $l'$ represent the vehicle's speed along the track direction and track tangential direction, respectively. These speeds are calculated as follows:

$$\dot{s} = v_x \frac{\cos \alpha}{1 - \kappa(s) \cdot l} \quad (14)$$

$$\dot{l} = (v_y \cdot \sin \alpha) \cdot \frac{1}{\dot{s}} \quad (15)$$

The action space, representing the control inputs that influence vehicle behavior, is defined as: $A = [\delta_f, T_{rt}, P_b]$. where $\delta_f$ refers to the vehicle's front wheel angle, serving as the control input that influences the vehicle's lateral dynamics. $T_{rt}$ denotes the rear-wheel-drive torque, which is the input for the longitudinal drive of a rear-wheel drive vehicle. $P_b$ represents the brake master cylinder fluid pressure.

For physical constraints, the limits of action variables $A = [\delta_f, T_{rt}, P_b]$ are determined by the vehicle's mechanical capabilities. Regarding safety constraints, the vehicle's safety state is represented by $\chi$. When a collision with the track boundary or a rollover occurs, $\chi = 0$, and the training episode is immediately terminated. For vehicle dynamics, the dynamic response used in our setup is provided by the *CarSim* platform, which offers a high-fidelity model of vehicle behavior.

*D. Iterative Optimization*

For the iterative optimization process, we selected the efficient Twin Delayed Deep Deterministic Policy Gradient (TD3) algorithm. TD3's twin-critic network architecture

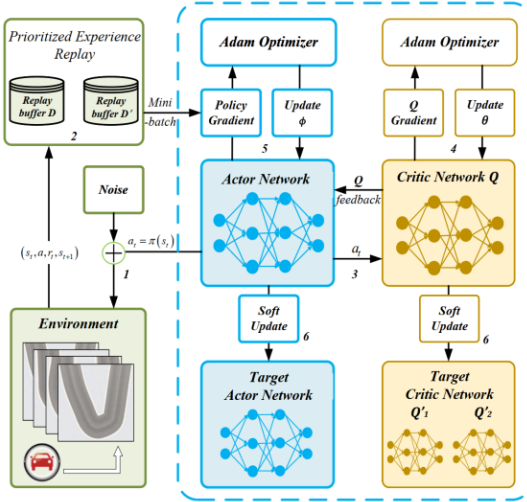

**Fig. 3.** Flowchart of Iterative Optimization of TD3 Algorithm

TABLE II THE IMPLEMENTATION PROCEDURE OF TD3

| Algorithm | Twin Delayed Deep Deterministic Policy Gradient (TD3) |
|---|---|

**Initialize randomly:** $Q(s,a;\theta_1)$ and $Q(s,a;\theta_2)$ policy function $\pi(a|s;\phi)$, an empty memory buffer $D$, and a relay buffer $D'$.

**For each episode do:**

   Obtain initial state $s_0$ following $\boldsymbol{d_{ini}(s)}$, Initialize a random $\boldsymbol{N}$

   **For each $t$** over $0,1,2,\ldots,N$ or until episode concludes

     1. $a_t \sim \pi(a|s_t;\phi) + n_t$

     2. Apply $a_i$, $D \leftarrow D \cup \{(s_t, a_t, r_t, s_{t+1})\}$, Sample a mini-batch

   **For each transition** do:

     3. Compute target value $y$: $a' \leftarrow \pi_{\phi'}(s') + clip(N', -c, c)$

$$y \leftarrow r_t + \gamma * \min_{k=1,2} Q_k(s', a'; \theta_k)$$

     4. Update critic networks:

$$\theta_k \leftarrow \theta_k - \lambda_q * \nabla_{\theta_k}(Q_k(s',a';\theta_k) - y)^2$$

   **If t mod policy_delay == 0** then:

     5. Update actor networks:

$$\phi \leftarrow \phi + \lambda_\pi * \nabla_\phi Q_1(s, \pi(a|s_t;\phi);\theta_1)$$

     6. Update target networks:

$$\theta_1' \leftarrow \tau * \theta_1 + (1-\tau) * \theta_1'; \quad \theta_2' \leftarrow \tau * \theta_2 + (1-\tau) * \theta_2'$$

$$\phi' \leftarrow \tau * \phi + (1-\tau) * \phi'$$

ensures accurate value estimations, while delayed policy updates and target policy smoothing enhance convergence and prevent the exploitation of narrow Q-function peaks [26]. Additionally, TD3 is a deterministic policy control strategy that provides consistent and reliable actions [27], ensuring trajectory stability in specific scenarios, which is crucial for the subsequent step of vehicle deployment. For policy iteration, TD3 maximizes a modified objective function to mitigate overestimation bias present in the traditional Deep Deterministic Policy Gradient (DDPG) approach. This is achieved by introducing twin Q-networks and a delayed policy update strategy. The objective function for TD3 is defined as follows:

$$Q_{target} = \min_{i=1,2} Q_{\theta_i}\left(s', \pi_{\phi'}(s')\right) \quad (16)$$

where $Q_{\theta_1}$ and $Q_{\theta_2}$ are the two critic networks used to estimate the Q-values, $\pi_{\phi'}$ is the target policy network, and $s'$ is the next state. The policy update occurs less frequently than the updates of the Q-networks to ensure policy evolution based on reliable value estimates, enhancing the learning process. The objective function for policy updates is expressed as:

$$J(\phi) = \mathbb{E}_{s \sim \mathcal{D}}\left[Q_{\theta_1}\left(s, \pi_\phi(s)\right)\right] \quad (17)$$

where $\mathcal{D}$ is the replay buffer from which samples are drawn. This formulation promotes a more stable and robust learning by systematically addressing the variance in policy updates and refining the exploration-exploitation balance. The combined strategy of using twin networks and delayed updates converges to an improved policy performance in complex decision-making environments.

Figure 3 and the pseudocode in the table II illustrate the iterative process of TD3. The Roman numerals in Figure 3 correspond to the steps in the pseudocode.

## IV. REAL VEHICLE DEPLOYMENT PROCESS

While directly adopting the TD3 actor policy performs well in simulation environments, it encounters challenges in real-world deployments, particularly in extreme cornering scenarios where highly nonlinear vehicle dynamics come into play.

Firstly, environmental variability poses a substantial obstacle. Unlike controlled simulation environments, real-world driving conditions are subject to unpredictable factors such as varying road surfaces, weather conditions, and dynamic obstacles. These factors introduce nonlinearities and uncertainties that are difficult to fully capture and model within simulations, leading to discrepancies between simulated and actual vehicle behavior.

Additionally, real-world sensors introduce noise and delays that are often neglected or idealized in simulations. These factors can adversely affect the controller's ability to make timely and accurate decisions.

Moreover, actuator constraints inherent to consumer-grade electric vehicles present significant challenges during deployment. These restrictions hinder the controller's ability to execute rapid and aggressive drift maneuvers effectively.

To address these, we propose a RL-MPC fusion controller that integrates the strengths of both RL and MPC. Figure 4 illustrates this real vehicle deployment scheme.

Essentially, both the RL controller and the MPC correction function are feedback controllers with a shared objective: high-speed cornering on the same curve. In real-world environments, the MPC compensates for the differences between the simulation environment and actual driving scenarios by tracking the optimal trajectory from the RL simulation. We believes that this integration enhances the controller's adaptability to varying environments, thereby preventing vehicle instability due to the simulation-to-reality gap.

### A. Preview Trajectory and Corrective Input Generation

Due to safety and computational power considerations, preview trajectory generation needs to be performed offline.

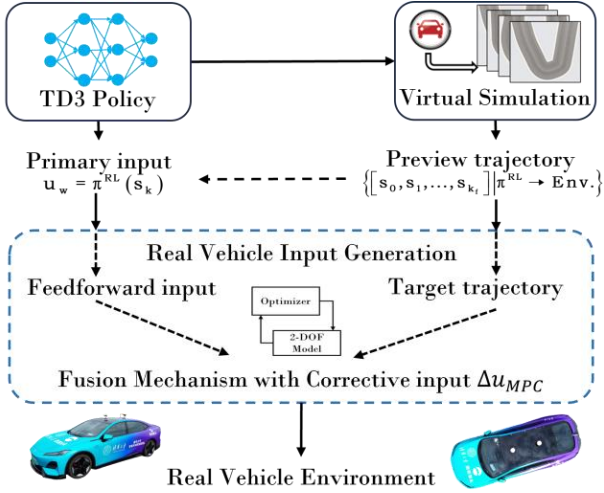

**Fig. 4**. Real Vehicle Deployment Scheme

This process begins by matching the virtual curve from the scene library with the actual curve environment. Once the appropriate virtual curve $V_c$ is selected, the current vehicle parameters $P_v$ are set. These parameters include the vehicle's weight and dynamic properties. The control policy $\pi$ obtained in Section III is then applied in the simulation environment using the TD3 policy to simulate the vehicle's behavior on the selected curve. This generates a preview trajectory $T_p$, which outlines the optimal path for the vehicle to follow. It is important to note that this preview trajectory differs from the pre-optimized trajectory used earlier in training.

$$T_p = \left\{ S_i \mid simulate(P_v, V_c, S_{ini}, \pi), i = 1, 2, 3, ..., n \right\} \quad (18)$$

where $S$ represents the vehicle state along the preview trajectory, as defined in Subsection III.C, and $S_{ini}$ denotes the vehicle state upon entering the curve. Note that since the preview trajectory is used to generate corrective input, $T_p$ needs to be converted into a Cartesian coordinate trajectory, $T_P^{xy} = [X, Y, \varphi, v_x, v_y, \dot\varphi]$. The conversion details are omitted here.

The RL inputs play a crucial role in enhancing the dynamic performance of the control system by predicting the global environment in advance [28]. The actions output by the RL policy are designed to reflect global optimality.

The RL inputs are determined by the TD3 policy $\pi$, which provides specific control actions corresponding to each state $S$ along the trajectory. For a given state $S$, the policy $\pi$ outputs an action $A_{RL}$ that serves as the primary input. This process can be mathematically represented as:

$$A_{ff} = \pi(S_t) = \left( \delta_f^{ff}, T_{rt}^{ff}, P_b^{ff} \right) \quad (19)$$

*B. Corrective Input Generation*

With appropriate primary inputs, the corrective inputs can compensate for the differences between the simulation environment and real driving scenarios, ensuring that the vehicle accurately tracks the target trajectory (i.e., the preview trajectory described in IV.A). Note that the trajectory used to generate the corrective input is **not** the pre-optimized trajectory obtained in Section III.A, but rather the trajectory produced by the RL agent during execution in the same simulation scenario.

To achieve this, we design a two-step forward model predictive controller (MPC) to generate the necessary corrective inputs. This controller adjusts the vehicle's actions in real-time, allowing it to precisely follow the preview trajectory while accounting for deviations or disturbances encountered during the drive.

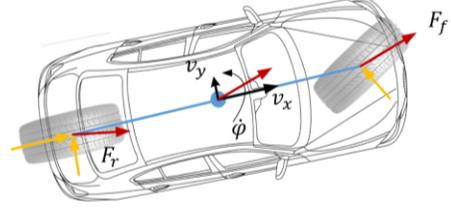

Fig. 5. Schematic diagram of the vehicle dynamics model used for MPC to generate corrective inputs.

First, we constructed a three-degree-of-freedom dynamic model of the vehicle for calculating the corrective input, as shown in Figure 5. All states are defined in a Cartesian coordinate system. To meet the real-time computation requirements, we chose a linear tire model [29] to calculate tire forces. Note that the dynamic response within the RL module is provided by the high-fidelity *Carsim* model, which is sufficient to generate control inputs for primary control, even in nonlinear conditions. The linear tire model, on the other hand, is used to compute a corrective input that addresses the differences between the simulated environment and real-world scenarios.

$$m\dot v_x = m v_y \dot\varphi + m a_{xt} \quad (20)$$

$$m\dot v_y = -m v_x \dot\varphi + 2\left[ C_{cf}\left( \frac{v_y + l_f \dot\varphi}{v_x} - \delta_f \right) + C_{cr} \frac{l_f \dot\varphi - v_y}{v_x} \right] \quad (21)$$

$$I_z \ddot\varphi = 2 l_f \left[ C_{cf}\left( \frac{v_y + l_f \dot\varphi}{v_x} - \delta_f \right) \right] - 2 l_r C_{cr} \frac{l_f \dot\varphi - v_y}{v_x} \quad (22)$$

In this dynamic model, $m$ represents the actual mass of the vehicle, and $I_z$ is the yaw inertia of the vehicle. In this model, $C_{cf}$ and $C_{cr}$ denote the cornering stiffness of the front and rear tires, respectively, while $l_f$ and $l_r$ denote the distances from the center of mass to the front and rear axles, respectively.

Converting the vehicle motion from the local body-fixed coordinate system to the inertial coordinate system involves the following transformation:

$$\dot X = v_x \cos\varphi - v_y \sin\varphi \quad (23)$$

$$\dot Y = v_x \sin\varphi + v_y \cos\varphi \quad (24)$$

At this point, we can establish the state equation 25 for generating corrective input using Equations 20-24. The six-state space variable vector can be denoted as $\Gamma = [X, Y, \varphi, v_x, v_y, \dot\varphi]$. The control variables are the steering angle and longitudinal acceleration, $u_t = [\delta_f, a_{xt}]$.

$$\dot\Gamma_t = A_t \Gamma_t + B_t \cdot u_t \quad (25)$$

The determination of $A_t$ and $B_t$ is well-established and will not be detailed here.

All variable information in $A_t$ and $B_t$ is included in the preview trajectory, so these variables can be considered known.

By substituting the current state information and the state information from the preview trajectory, the state equation becomes linear.

Before the MPC controller calculates the correction input, the continuous response equations should be discretized. Also, to account for input change rate constraints due to the mechanical structure, new state-space equations are constructed.

$$\tilde{\Gamma}(k+1) = f\left(\tilde{\Gamma}(k), \tilde{u}(k)\right) = \tilde{A}_k \tilde{\Gamma}(k) + \tilde{B}_k \tilde{u}(k) \quad (26)$$

$$\tilde{\Gamma}(k) = \begin{bmatrix} \Gamma(k)^T & u(k-1)^T \end{bmatrix}^T, \quad \tilde{u}(k) = u(k) - u(k-1) \quad (27)$$

$$A_k = \begin{bmatrix} e^{A_t(\Gamma_t)T_s} & \int_0^{T_s} e^{A_t(\Gamma_t)\tau} d\tau \cdot B_t \\ 0_{2\times 6} & I \end{bmatrix} \quad (28)$$

$$B_k = \begin{bmatrix} \int_0^{T_s} e^{A_t(\Gamma_t)\tau} d\tau \cdot B_t \\ I \end{bmatrix} \quad (29)$$

The prediction and control horizons of the controller are both set to two steps ahead.

$$\tilde{\Gamma}(k+2) = \tilde{A}_{k+1}\tilde{A}_k \tilde{\Gamma}(k) + \tilde{A}_k \tilde{B}_k \tilde{u}(k) + \tilde{B}_{k+1}\tilde{u}(k+1) \quad (30)$$

$$\tilde{\Gamma}_f(k) = \begin{bmatrix} \tilde{\Gamma}(k+1) & \tilde{\Gamma}(k+2) \end{bmatrix}^T, \quad \tilde{u}_f(k) = \begin{bmatrix} \tilde{u}(k) & \tilde{u}(k+1) \end{bmatrix}^T \quad (31)$$

To minimize the trajectory error during tracking process, the following cost function is constructed:

$$J_k = \sum_{t=k+1}^{k+2} \left( \left\| \Gamma(t) - T_p^{xy} \right\|_Q^2 - \left\| \Delta \tilde{u}(t-1) \right\|_R^2 \right) \quad (32)$$

The weight matrices $Q$ and $R$ are used to penalize state errors, the rate of change in control inputs, respectively.

At this point, the input corresponding to each time step $k$ can be obtained by solving the following optimization problem:

$$\begin{aligned}
&\min_{\tilde{u}_k} J_k \\
&s.t. \quad \tilde{\Gamma}(k+1) = f\left(\tilde{\Gamma}(k), \tilde{u}(k)\right) \\
&\quad\quad u(k) = u(k-1) + \tilde{u}(k) \\
&s.t. \quad \tilde{u}_{\min} \leq \tilde{u}(k) \leq \tilde{u}_{\max} \\
&\quad\quad u_{\min} \leq u(k) \leq u_{\max}
\end{aligned} \quad (33)$$

This optimization problem can be solved using Quadratic Programming (QP) to obtain the control input $u_k$. The QP approach is well-established and efficient. Given its maturity, the details of the solution process are not elaborated here. The acceleration $a_{xt}$ is converted into the actual input $T_{rt}, P_b$ through the built-in tracking logic.

At this stage, the MPC generates corrective inputs $\Delta u_{MPC} = u_k$ based on the current task's objectives, which are used to compensate for the differences between the simulation environment and actual driving scenarios. These corrective inputs are combined with the primary RL input $A_{RL}$ to adjust and refine the overall control actions. The final input $u_t$ is expressed as:

$$u_t = A_{RL} + \Delta u_{MPC} \quad (34)$$

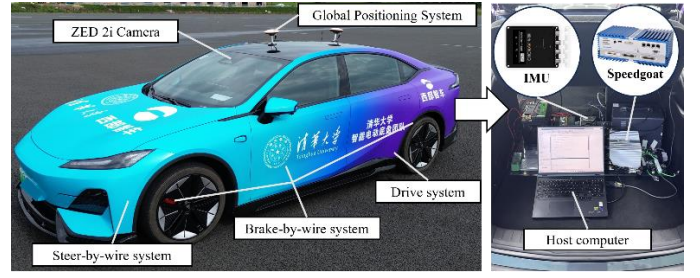

Fig. 6. The comprehensive setup of the real vehicle test platform.

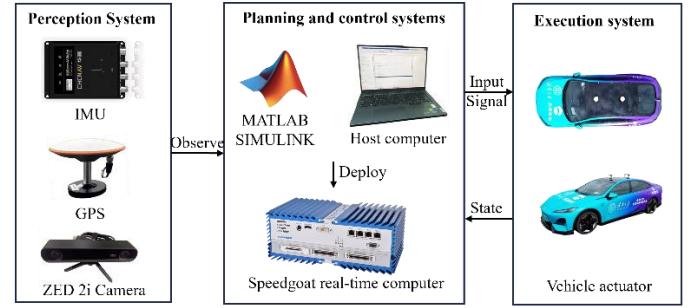

Fig. 7. The control flow of the real vehicle platform.

This integrated RL-MPC fusion mechanism provides robust control for the vehicle, ensuring it follows the optimal trajectory in real-world conditions.

Although a key goal is to avoid vehicle instability due to environmental differences, instability can still occur under extreme conditions. This usually presents as excessive oversteer, where the vehicle's rear swings out, causing loss of control. To manage this, we have a safety fallback: when the side-slip angle exceeds a threshold, moderate braking replaces the controller's drive-brake input: $T_{rt} = 0, P_b = P_{bm}$.

## V. REAL VEHICLE TESTS

The core of the previous work is to deploy the RL-based controller on a real vehicle in a real traffic environment. This section introduces the consumer electric car used for verification and demonstrates the effectiveness of the method in executing drift U-turns and drift right-angle turns.

### A. Real-Vehicle Test Platform

The verification platform for this study is a full-scale consumer-grade rear-wheel drive electric vehicle, as shown in Figure 6. The vehicle features a steer-by-wire electric power steering system, allowing for closed-loop control of the front wheel angle. The drive system provides closed-loop control of the torque, and the brake-by-wire hydraulic braking system enables closed-loop control of the master cylinder pressure. The platform is also equipped with cameras, inertial navigation, a global positioning system, and a fast real-time industrial computer (*Speedgoat*). The control flow of the real vehicle platform is shown in Figure 7. The vehicle motion state observation signals collected by sensors, such as the CGI 610, GNSS, include vehicle position, longitudinal and lateral speed, acceleration, yaw rate, wheel speed. These signals are transmitted to the real-time machine *Speedgoat*. The control algorithm is developed and implemented in the MATLAB/

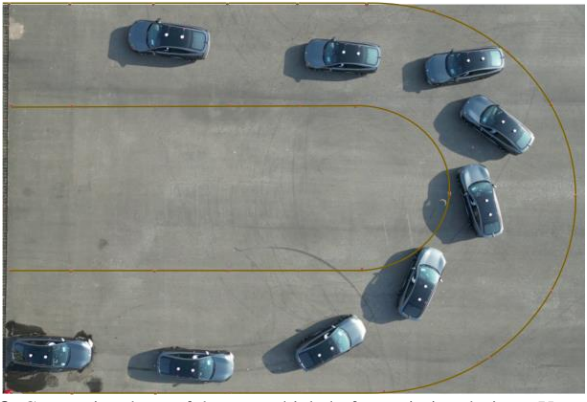

**Fig. 8:** Composite photo of the test vehicle before painting during a U-turn maneuver (top view) – each frame was taken 0.7 seconds apart (except for the first and final intervals). Photo credit: Shiyue Zhao.

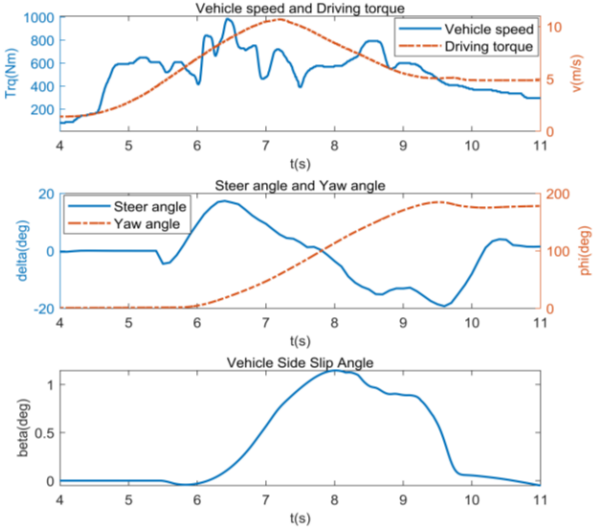

**Fig. 9:** Key vehicle states and inputs during the U-turn maneuver.

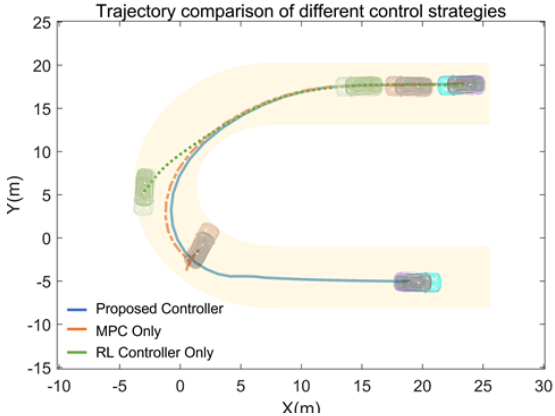

**Fig. 10:** Trajectory comparison of different control strategies.

Simulink environment on a host computer, then compiled and deployed to the *Speedgoat* for real-time operation. The *Speedgoat* is a portable real-time machine equipped with an Intel i7 2.5 GHz dual-core CPU and supports CAN FD communication.

Based on the complexity of the vehicle state and control algorithm, a control task frequency of 10 ms (100 Hz) was selected. The team's iterative adjustments and trials during real vehicle testing demonstrated that this frequency is sufficient to capture rapid changes in vehicle dynamics, ensuring real-time controller adjustments while providing a significant computational margin for task execution.

### B. U-turn Extreme Cornering Test

To verify the effectiveness of the proposed control strategy, a U-turn extreme cornering test was conducted. The vehicle was subjected to a series of drift U-turn maneuvers to evaluate the performance of the proposed RL-MPC fusion control.

From a functional perspective, the actual vehicle test results demonstrate that this research method can autonomously control a vehicle to complete the U-turn task on a U-turn curve, as shown in Figure 8. A video of the entire process can be viewed through the link provided in the appendix. During the test, tire smoke and strong friction noise were observed, indicating the extreme nature of the cornering.

From a performance perspective, when negotiating an extreme center radius of 11m and a width of 5.5m, the time from entering to exiting the corner was 4.8s. Figure 9 presents the key states and inputs of the vehicle during this maneuver. The upper section displays the vehicle speed and wheel-end drive torque, the middle section shows the front wheel angle and yaw rate, and the lower section illustrates the vehicle's center of mass side-slip angle, a critical indicator of extreme driving conditions.

Speed analysis reveals that the proposed controller achieved a maximum cornering speed of 10.4 m/s. The minimum speed during the entire trajectory was not less than 4.4 m/s, indicating that the vehicle maintained a considerable speed even in highly nonlinear dynamic areas, which quantitatively demonstrates extreme cornering. The center of mass side-slip angle analysis shows a maximum side slip angle of 63.7 degrees. From 7.5s to 9.3s, the side-slip angle remained around 55.3 degrees, indicating an extreme motion phenomenon close to a steady state. Notably, after 7.9s, the direction of the center of mass side-slip angle was opposite to the front wheel angle and was maintained for a period, highlighting the highly nonlinear dynamic response of the vehicle. This demonstrates that the proposed controller can effectively manage the vehicle in highly nonlinear dynamic conditions.

For comparison, we selected two strategies: the first used the RL controller proposed in Section III, which achieved satisfactory control effects in simulation; the second used only the MPC controller proposed in literature 29 to track the target preview trajectory, which has been proven effective for various cornering trajectories. Figure 10 compares the trajectories of the proposed strategy and the two comparative strategies. The blue line represents the proposed strategy, the green line represents the RL controller alone, and the orange line represents the MPC controller alone. Table III presents evaluation indicators for the three strategies and the RL strategy's control effect in simulation. The total time refers to the overall time spent driving in the roundabout.

The trajectory executed by the RL controller collided with the track boundary soon after entering the drift state, likely due to the gap between the real and simulated environments. The analysis shows that the most significant interference factor arises from changes in the tire's ground adhesion coefficient, especially when the tire slips and experiences sharp temperature changes. The MPC strategy could not stabilize the vehicle after entering the highly nonlinear area, causing excessive drifting and stopping, likely due to the MPC's reliance on a linear vehicle dynamics model. Among the three strategies, only the proposed strategy successfully completed the extreme U-turn task, demonstrating a functional innovation that ensures safety without sacrificing performance.

TABLE III EVALUATION INDICATORS FOR CONTROL POLICIES

| Policy | Task Completion (deg) | Max. speed (m/s) | Max. side-slip angle (deg) | Total time (s) |
|---|---|---|---|---|
| RL policy in simulators | 180/180 | 11.0 | 67.4 | 4.52 |
| RL policy In real world | 101/180 | 12.7 | 37.4 | -- |
| MPC policy In [30] | 142/180 | 10.9 | 83.8 | -- |
| **Proposed policy** | **180/180** | **10.4** | **63.7** | **4.81** |

*C. Right-angle Turns Extreme Cornering Test*

To further verify the effectiveness and generalization of the proposed control strategy, a right-angle turn limit test was conducted. In this experiment, tests were performed on a wet, medium adhesion road surface after rain to assess its effectiveness under varying road conditions.

From a functional perspective, the actual vehicle test results demonstrate that the proposed method can autonomously control the vehicle to complete the turning task on a right-angle curve. The test was accompanied by strong tire friction sounds, indicating the extremity of the turning process.

From a performance perspective, when navigating a right-angle curve with a center radius of 11m and a width of 5.5m, the time from entering to exiting the curve was 3.7 seconds. Figure 11 illustrates the key states of the vehicle during this maneuver. Although the turning angle in this scenario is small, the vehicle experiences a continuously changing dynamic drift state during control, demonstrating that the controller achieves good dynamic control effects. Other state analyses yield conclusions consistent with those of the U-turn maneuver and will not be expanded upon here. We also employed the same comparative strategies described in IV.B for this curve, with the comparison results shown in Table IV. Due to space constraints, the trajectory diagram is not included. Figure 12 presents the RL Input and Corrective Input for Steering and Driving Torque. It demonstrates the controller's ability to adapt to discrepancies in tire characteristics and environmental disturbances between the simulation environment and real-world deployment.

The results indicate that the RL policy in simulators achieves high task completion with fast speed and moderate side-slip angle. However, in real-world conditions, the RL policy struggles due to environmental differences. The MPC policy in [30] completes the task but requires more time, showing a conservative approach that prioritizes stability over speed. The proposed policy effectively balances speed and control, achieving complete task performance with a high maximum speed and controlled side-slip angle in a reasonable time.

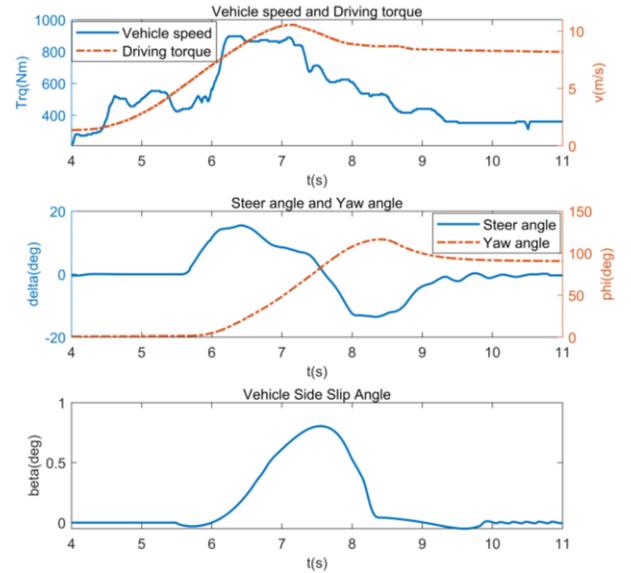

Fig. 11: Key vehicle states and inputs during right-angle turns extreme cornering.

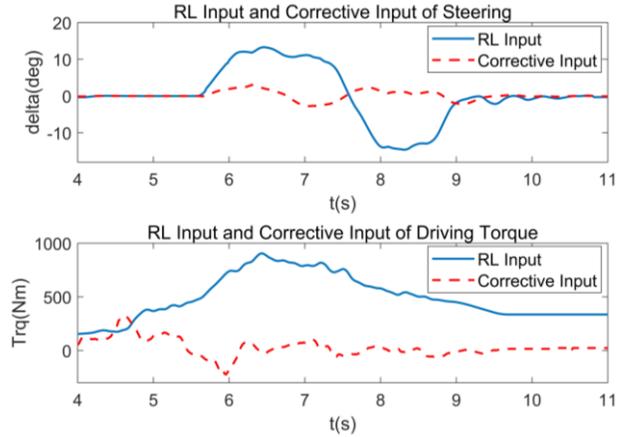

Fig. 12: RL Input and Corrective Input Comparison for Steering and Driving Torque

TABLE IV EVALUATION INDICATORS FOR CONTROL POLICIES

| Policy | Task Completion (deg) | Max. speed (m/s) | Max. side-slip angle (deg) | Total time (s) |
|---|---|---|---|---|
| RL policy in simulators | 90/90 | 11.2 | 52.6 | 2.81 |
| RL policy In real world | 47/90 | 11.8 | 61.7 | -- |
| MPC policy In [30] | 90/90 | 10.9 | 50.1 | 6.43 |
| **Proposed policy** | **90/90** | **10.6** | **46.1** | **3.19** |

To evaluate computational real-time performance, we analyzed the logs of the *Speedgoat* real-time machine while executing the proposed control method. Within each task execution cycle, the average computation time was 3.35 ms, with a minimum of 2.93 ms and a maximum of 7.84 ms. These results further validate the previously established control frequency as appropriate.

*D. Robustness Analysis of Proposed Controller*

For the influence of real-world parameters such as road conditions on the effectiveness of the proposed control approach, we conducted additional experiments to evaluate the controller's robustness to variations in tire-road adhesion coefficients, which are critical for drift maneuvers. The peak adhesion coefficient at our test site was measured to be approximately between 0.8 and 0.9. Due to experimental constraints, we could not adjust or precisely control the adhesion coefficient of the actual test surface. Therefore, we varied the adhesion coefficient within the RL training environment.

By altering this parameter during deployment, we observed the controller's performance under different simulated friction levels, directly reflecting the extent to which our controller can adapt to simulation-to-reality discrepancies. The primary evaluation metrics were the completion of the cornering task and the cornering time. The results are summarized in Table V.

TABLE V CONTROLLER PERFORMANCE UNDER VARYING ADHESION COEFFICIENTS IN TRAINING ENVIRONMENT

| Adhesion coefficient in training environment | U-turn task and time (s) | Right-angle turn task and time (s) |
|---|---|---|
| 0.95 | 4.86 | 3.26 |
| 0.85 | 4.81 | 3.19 |
| 0.75 | 5.16 | 3.98 |
| 0.65 | 7.24 | 4.66 |
| 0.55 | N/A (127/180) | N/A (53/90) |

The results demonstrate that the proposed control method successfully completed the cornering tasks with adhesion coefficients in the training environment ranging from 0.95 to 0.65, with only a gradual increase in cornering time as adhesion decreased. This trend indicates that the controller adapts to reduced friction by sacrificing some performance; although the time increases, the task completion is still maintained. However, at an adhesion coefficient of 0.55, the controller was only able to partially complete the maneuvers, achieving 127 degrees out of 180 degrees for the U-turn and 53 degrees out of 90 degrees for the right-angle turn. This indicates that when the friction coefficient deviation between the training environment and the actual application environment exceeds a certain threshold, the controller's ability to execute drift maneuvers is significantly affected due to insufficient ground force to support maneuvers that could be completed in the simulation environment.

Notably, directly deploying an RL policy trained in an environment as similar as possible to real-world conditions still resulted in failures for high-speed drift maneuvers, as shown in Tables III and IV. This is because scenario-to-action RL agents rely on precise vehicle dynamics and road environment models, which are challenging to accurately represent in simulations. In contrast, our proposed real vehicle deployment method not only successfully completes the cornering tasks but also tolerates a certain degree of modeling deviations. These findings confirm the robustness of our control approach to variations in tire-road adhesion coefficients, thereby enhancing the framework's feasibility for real-world applications.

## VI DISCUSSION

This study presents the first deployment of a RL-based transient drift algorithm on real consumer-grade electric vehicles, significantly enhancing their handling during high-speed cornering and enabling effective high-speed drift maneuvers. The proposed method effectively controls vehicles during U-turn and right-angle turn maneuvers, demonstrating robust handling in highly nonlinear dynamic states. It maintains high speeds and significant side-slip angles during extreme maneuvers, showing superior dynamic control. Compared to traditional RL and MPC strategies, the proposed controller offers better stability and trajectory accuracy, effectively bridging the gap between simulation and real-world applications. These results confirm the efficacy of the proposed strategy in managing complex driving scenarios, enhancing both safety and performance for autonomous vehicle deployment. Furthermore, by effectively bridging simulation-to-reality gaps, we provide a viable solution for deploying scenario-to-action RL controllers that handle strongly nonlinear vehicle dynamics in real-world autonomous driving environments. The successful real-vehicle tests underscore the potential of our control framework to advance autonomous driving technologies, offering enhanced handling capabilities and robust performance across diverse and unpredictable driving conditions.

However, actuator constraints significantly limit our control framework's performance on consumer-grade electric vehicles. Limitations such as a front wheel steering rate capped at -43.65°/s, peak drive torque of 1000 Nm, and drive torque filtering hinder rapid drift maneuvers. Additionally, discrepancies between simulated and actual actuators require precise calibration, directly impacting drift cornering results. Future work will address these constraints by adopting higher-performance hardware or enhancing actuator modeling in simulations, thereby improving the controller's effectiveness and broadening its applicability to various vehicle types and driving scenarios.

## VII. APPENDIX

The videos provide a visual demonstration of the drift turning tests. To view the videos, please see the supplementary materials or visit the link: https://youtu.be/5wp67FcpfL8.

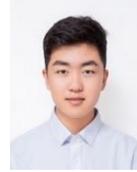

**Shiyue Zhao** is currently a visiting scholar at the University of Michigan, Ann Arbor, USA. He received the B.S. degree in traffic engineering from Central South University, Hunan, China, in 2021, and is pursuing the Ph.D. degree at the School of Vehicle and Mobility, Tsinghua University, Beijing, China. His research interests include transportation systems, advanced vehicle control, and vehicle extreme control.

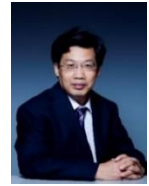

**Junzhi Zhang** received the B.S., M.S., and Ph.D. degrees in transportation from Jilin University, Changchun, China, in 1992, 1995, and 1997, respectively. He is currently a Professor in Power Engineering and Engineering Thermophysics with the State Key Laboratory of Automotive Safety and Energy, Tsinghua University, Beijing, China.

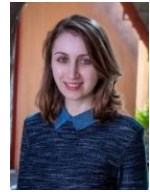

**Neda Masoud** received her B.S. in Industrial Engineering from Sharif University of Technology, M.S. in Physics from the University of Massachusetts, and Ph.D. in Civil Engineering from the University of California, Irvine. She is currently an Associate Professor in Civil and Environmental Engineering at the University of Michigan, Ann Arbor.

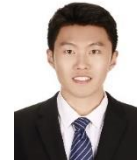

**Yuhong Jiang** received the B.E.degree from School of Vehicle and Mobility, Tsinghua University in 2023. He is currently studying as a Ph.D. student at the same institution. His research interests include vehicle dynamics control and extreme motion control for autonomous vehicles.

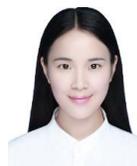

**Heye Huang** received the B.E. degree from Central South University, Changsha, in 2018, and the Ph.D. degree from Tsinghua University in 2023. She is currently a Research Associate Fellow with Connected & Autonomous Transportation Systems Lab, University of Wisconsin–Madison. Her current research interests include connected and automated vehicles, risk assessment, decision making and motion prediction.

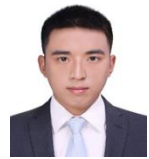

**Tao Liu** received his bachelor's degree in automotive engineering from Tongji University in 2012. He is currently studying for a master's degree in engineering management at Tsinghua University. He has been deeply engaged in automotive technology research and development at Ford Motor Company from 2017 to 2022 and has achieved a number of awards.